\documentclass[conference,10pt]{IEEEtran}
\IEEEoverridecommandlockouts 
\usepackage{blkarray}                                      
\usepackage{algpseudocode}                                 
\usepackage{algorithm}
\usepackage{graphicx}                                      
\usepackage{amsmath}
\usepackage{amssymb}
\usepackage{amsfonts}
\usepackage{amsthm}
\usepackage[mathcal]{eucal}
\usepackage{mathrsfs}
\usepackage{booktabs}
\usepackage{enumerate}
\usepackage{multirow}
\usepackage[subrefformat=parens,farskip=0pt,justification=centering]{subfig}
\usepackage{color}
\usepackage{cite}                                          
\usepackage{comment}                                       
\usepackage{soul}                                          
\soulregister\cite7
\soulregister\ref7
\soulregister\pageref7
\usepackage{etoolbox}                                      
\usepackage{url}
\usepackage{nth}                                           
\usepackage{bm}                                            
\usepackage{courier}
\usepackage{balance}
\usepackage{threeparttable}
\usepackage[bookmarks=false]{hyperref}
\hypersetup{
    colorlinks = true,
    citecolor  = blue,
    linkcolor  = blue,
    urlcolor   = blue,
}
\usepackage{tikz}
\usetikzlibrary{positioning,calc,fit,decorations.pathmorphing}
\usepackage{filecontents}                                  
\usepackage{pgfplots}
\usepackage{pgfplotstable}
\pgfplotsset{compat=newest}
\usepackage{caption}
\usepackage{cleveref}
\Crefformat{figure}{Fig.~#2#1#3}                           
\Crefname{subfigure}{Fig.}{Figs.}
\Crefname{figure}{Fig.}{Figs.}
\Crefformat{table}{TABLE~#2#1#3}                           
\definecolor{CUHKorange}{RGB}{244,106,18} 
\definecolor{CUHKblue}{RGB}{0,111,190}    
\definecolor{CUHKgreen}{RGB}{0,127,128}   
\definecolor{CUHKred}{RGB}{228,46,36}     
\definecolor{CUHKyellow}{RGB}{198,148,34} 
\definecolor{CUHKdark}{RGB}{114,44,114}   
\definecolor{CUHKmiddle}{RGB}{144,44,144} 
\definecolor{CUHKlight}{RGB}{167,44,167} 

\newcommand{\subparagraph}{}
\usepackage{titlesec}
\titlespacing{\section}{0pt}{1.6ex plus .2ex minus .2ex}{0.4ex plus .1ex}
\titlespacing{\subsection}{0pt}{1.0ex plus .2ex minus .2ex}{0.3ex plus .1ex}

\newtheorem{myproblem}{\textbf{Problem}}
\newtheorem{mydefinition}{\textbf{Definition}}

\algrenewcommand\textproc{\texttt}

\makeatletter
\let\OldStatex\Statex
\renewcommand{\Statex}[1][3]{%
  \setlength\@tempdima{\algorithmicindent}%
  \OldStatex\hskip\dimexpr#1\@tempdima\relax
}
\makeatother

\RequirePackage[normalem]{ulem} 
\RequirePackage{color}\definecolor{RED}{rgb}{1,0,0}\definecolor{BLUE}{rgb}{0,0,1} 

\definecolor{myblue}{RGB}{29,114,221}    
\definecolor{myyellow}{RGB}{255,255,191} 
\definecolor{myorange}{RGB}{244,106,18}  
\definecolor{mygray}{RGB}{102,102,102}   
\definecolor{mygray2}{RGB}{166,166,166}  %
\definecolor{mypink}{RGB}{252,228,215}   

\graphicspath{{./figs/}}
\usepackage{pgfplots}
\usepackage{multirow}
\usepackage{makecell}
\usepackage{threeparttable}
\usepackage{pifont}

\definecolor{myorange}{RGB}{244,106,18} 
\definecolor{myblue}{RGB}{0,111,190}    
\begin{document}
\date{}

\title{
    LMM-IR: \underline{L}arge-Scale Netlist-Aware \underline{M}ulti\underline{m}odal Framework for Static \underline{IR}-Drop Prediction
}

\author{
       \IEEEauthorblockN{
          Kai Ma$^1$,
          Zhen Wang$^1$,
          Hongquan He$^1$,
          Qi Xu$^2$,      
          Tinghuan Chen$^3$,
    	 Hao Geng$^{1{\dagger}}$
   }
    \IEEEauthorblockA{
	 $^1$ShanghaiTech University \quad
      $^2$University of Science and Technology of China \\
      $^3$The Chinese University of Hong Kong, Shenzhen
	 }
}

\makeatletter
\renewcommand{\footnoterule}{%
  \kern-3pt
  \hrule\@width 1in 
  \@height 0.4pt 
  \kern 2.6pt
}
\makeatother
\maketitle

\begin{abstract}
Static IR drop analysis is a fundamental and critical task in the field of chip design. Nevertheless, this process can be quite time-consuming, potentially requiring several hours. Moreover, addressing IR drop violations frequently demands iterative analysis, thereby causing the computational burden. Therefore, fast and accurate IR drop prediction is vital for reducing the overall time invested in chip design.
In this paper, we firstly propose a novel multimodal approach that efficiently processes SPICE files through large-scale netlist transformer (LNT). Our key innovation is representing and processing netlist topology as 3D point cloud representations, enabling efficient handling of netlist with up to hundreds of thousands to millions nodes. All types of data, including netlist files and image data, are encoded into latent space as features and fed into the model for static voltage drop prediction. This enables the integration of data from multiple modalities for complementary predictions.
Experimental results demonstrate that our proposed algorithm can achieve the best F1 score and the lowest MAE among the winning teams of the ICCAD 2023 contest and the state-of-the-art algorithms.
\end{abstract}


\def\thefootnote{$\dagger$}\footnotetext{Corresponding author.}\def\thefootnote{\arabic{footnote}}
\section{Introduction}
\label{sec:introduction}

\begin{figure}[tb!]
    \centering    \includegraphics[width=1.0\linewidth]{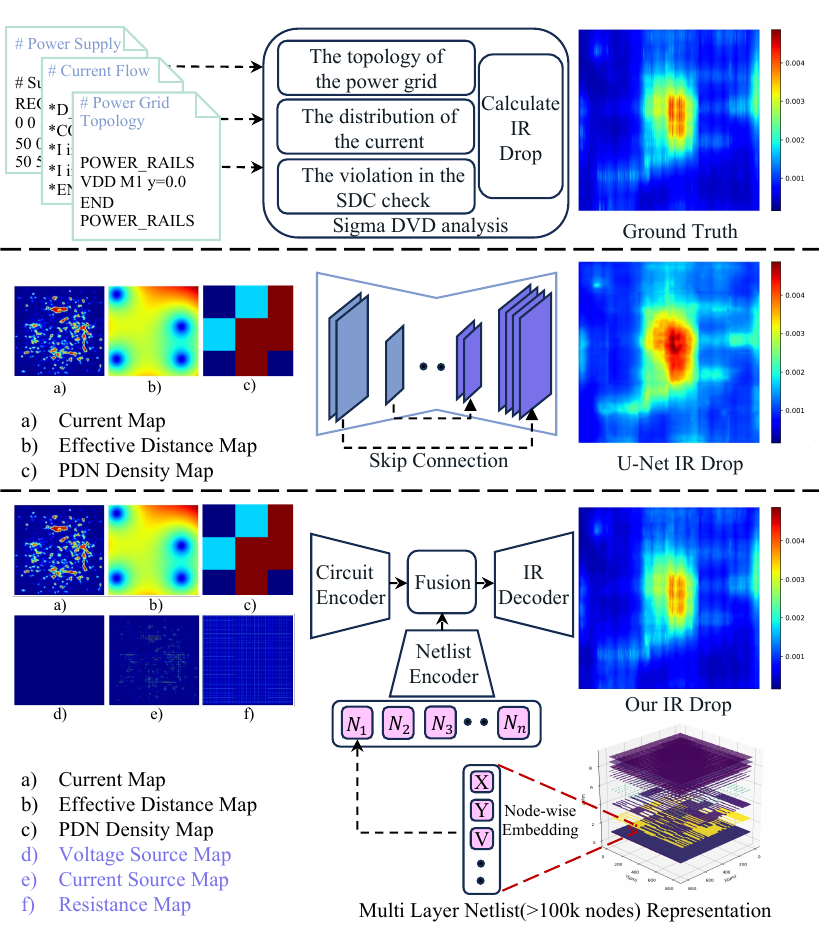}
    \caption{The workflow of three models for IR drop prediction. From top to bottom: commercial tool, traditional U-Net, and the proposed LMM-IR model.
    }
    \label{fig:brief_flow}
\end{figure}

As AI and high-performance computing (HPC) advance in tandem, new pathways like chiplet \cite{li2020chiplet} bring design challenges and significant power consumption concerns. The analysis and minimization of integrity losses within signal and power distribution networks (PDNs) have become increasingly crucial. 
Each transistor, whether nMOS or pMOS, requires power, which is distributed across the microprocessor by metal layers \cite{power-1}. While chip dimensions have remained relatively constant, advancements in fabrication technology have led to thinner, though not shorter, wires. This trend has resulted in a significant increase in resistance per unit length. For instance, between $ 28 $ nm and $ 7 $ nm nodes, there is nearly a tenfold increase in resistance, with exponential growth expected for smaller geometries. 
Voltage drops when current passes through a resistor. 
A decline in voltage can result in a reduction in the speed of a transistor, which may potentially lead to a functional failure if it occurs on a critical path within a design \cite{power-2}. 
The potential for performance degradation and functional failures requires a comprehensive approach to IR drop analysis throughout the design cycle. 
IR drop stands as a pivotal focus for numerous EDA vendors. 

To address these challenges, design engineers must repeatedly assess and address IR drop issues across the various design phases, extending from the initial placement to the final sign-off. The workflow of the commercial tool is shown in \Cref{fig:brief_flow}. However, the use of simulation-based methods to acquire precise IR drop estimates is an extremely time-consuming process \cite{time-consuming}. Consequently, IR drop mitigation strategies that rely on frequent simulations usually require a large amount of computational resources and time, especially when performing static voltage drop simulations on a chip-wide scale. To enhance efficiency in this process, there is a crucial demand for an IR drop estimation method that is both fast and accurate.
With the rapid advancement of artificial intelligence (AI), an increasing number of machine learning (ML)-based methods have been proposed for predicting static IR drop.
PowerNet \cite{powernet} introduced a cell-by-cell prediction methodology utilizing convolutional neural networks (CNNs). However, it's challenging to find the right cell size, and processing each cell individually is time-consuming. Besides, IREDGe \cite{thermal} employed the U-Net \cite{unet} architecture to analyze the IR-drop while the conventional CNNs often fall short in capturing comprehensive global circuit context, leading to potential inaccuracies. Additionally, recent advancements such as IRPnet \cite{date24} proposed a shape-adaptive conventional kernel to capture more information from the physical structure. Nevertheless, the methodology employed by IRPnet is constrained by its dependence on single-modality features, which may be insufficient for capturing the intricate characteristics inherent in circuits. 
To address these fundamental challenges, we propose LMM-IR, a novel multi-modal fusion architecture. Our framework introduces an end-to-end paradigm with GPU acceleration that autonomously optimizes IR drop predictions across the entire chip. The LMM-IR architecture innovatively integrates a large-scale netlist transformer (LNT) to comprehensively represent netlist features, enabling superior capture of three-dimensional spatial relationships among metal layers. By aligning and fusing netlist data with circuit-level information, our approach facilitates mutual enhancement of diverse modalities, providing a more robust and accurate representation for IR-drop prediction.
The main differences between our model, called LMM-IR, and previous machine learning-based works are summarized in \Cref{tab:difftech}.

Departing from conventional simulation-based techniques and single modality model, we propose a novel large-scale netlist-aware multimodal framework that fundamentally transforms static voltage drop estimation, delivering superior accuracy and computational efficiency. An illustration of our multimodal framework is shown in \Cref{fig:brief_flow}.

Our contributions are summarized as follows:
\begin{itemize}
    \item To the best of our knowledge, the first end-to-end framework is proposed which can directly process large-scale netlist data (more than 100K nodes) alongside circuit data, achieving state-of-the-art performance while maintaining competitive inference time. This breakthrough bridges the gap between netlist-level and circuit-level analysis in IR drop prediction.

    \item A novel large-scale netlist transformer (LNT) is designed for netlist feature extraction. This innovation enables comprehensive modeling of inter-layer connections and significantly enhances the model's ability to understand chip topology.

    \item For the first time, we propose a novel multi-modal fusion architecture that effectively aligns and integrates netlist and circuit data, facilitating mutual enhancement of different modalities.

    \item Extensive experiments demonstrate our approach's performance on large and complex circuits, overcoming the limitations of traditional image-based methods. The multimodal approach yields superior results.
\end{itemize}

\newcommand{\cmark}{\ding{51}}
\newcommand{\xmark}{\ding{55}}

\begin{table}[tb!]
    \centering
    \caption{Comparison among different IR drop models.}
    \begin{threeparttable}
        \setlength{\tabcolsep}{1.65mm}
        {
            \begin{tabular}{c|c|c|c|c}
                \toprule
                Methods & \makecell{Fully handle\\Netlist} & \makecell{Multimodal\\Fusion} & \makecell{Extra\\Features} & \makecell{Global attention\\mechanism} \\
                \midrule
                1st Place\tnote{+} \cite{iccad23} & \xmark & \xmark & \cmark & \cmark \\
                2nd Place\tnote{+} \cite{iccad23} & \xmark & \xmark & \cmark & \cmark \\
                IREDGe \cite{began} & \xmark & \xmark & \xmark & \xmark \\
                IRPnet \cite{date24} & \xmark & \xmark & \xmark & \xmark \\
                \midrule
                Ours & \cmark & \cmark & \cmark & \cmark \\
                \bottomrule
            \end{tabular}
            \begin{tablenotes}
                \footnotesize
                \item[+] ``1st and 2nd Place'' refer to the top two winners of 2023 ICCAD contest.
            \end{tablenotes}
        }
    \end{threeparttable}
    \label{tab:difftech}
\end{table}
\section{Preliminaries}
\label{sec:prelim}

\subsection{Circuits Representation}
\label{sec:circuit}
The on-chip PDN structure \cite{samal2016full} can be modeled as a network consisting of \textit{voltage sources}, \textit{current sources}, and \textit{resistors}, where the wires form the resistive network, the power source is a voltage source connected to the wires of the PDN and the current source represents the cell/instance consuming the current. 
The goal of the static voltage drop simulation is to determine the voltage at the nodes in the PDN to which the instances (current sources) are connected.

The input data is provided in 2 formats:

1). \textit{Circuit-based data:} There are $3$ inputs determining the static IR drop, including current map, effective distance to the voltage source, and PDN density map. 
The circuit-based data is provided as matrices in comma-separated values (CSV) files, with each value in the file representing the current, effective distance, or PDN density in a $1 \mu m \times 1 \mu m$ region of the chip (noting that the spacing of the PDN nodes at the lowest level will be larger). 

2). \textit{SPICE-based data:} For each data point, the PDN model is provided as a SPICE netlist. The SPICE netlist describes the node positions, resistance values between nodes, current source locations and their values, as well as voltage source location and their values. 

\subsection{Multimodal Representation Learning}
The field of multimodal representation learning\cite{10123038} has attracted significant interest due to its capacity to integrate diverse data sources and enhance predictive performance in complex systems. This approach is particularly relevant in the field of electronic design automation (EDA), where integrating heterogeneous data types, including circuit-level characteristics and netlist descriptions, is essential for improving the accuracy of tasks such as IR drop prediction. Alternative methodologies, such as CLIP \cite{radford2021clip} and SLIP \cite{mu2022slip}, generate distinct features for images and text independently, followed by an alignment phase. This efficient process enables large-scale pre-training while remaining robust in the presence of noisy or incomplete data. In electronic design automation (EDA), multimodal learning facilitates the combination of graphical circuit representations with structured netlist data, providing richer feature sets that enhance modeling precision and predictive capability. By integrating data from different modalities within a unified framework, this approach strengthens system analysis, particularly for critical applications such as power integrity and signal integrity verification.

\subsection{Attention Module in IR Drop Prediction}

Special attention must be paid to the prediction of hot areas, as their scope of hot areas is too small and their proportion in the whole network is not large. We employ the attention module from the Transformer\cite{vaswani2017attention} and modify it to incorporate cross-attention for the multimodality fusion and self-attention for feature extraction within the netlist embedding. Let \textit{Q}, \textit{K}, and \textit{V} represent the $query$, $key$, and $value$, respectively. Given an input $\textit{X}\in\mathbb{R}^{n\times c}$ where $n$ is the count of embedding entities and $c$ is the number of channels per entity. Consider three weights $\textit{W}_Q, \textit{W}_K \in \mathbb{R}^{c\times d_\alpha}$ and $\textit{W}_V \in \mathbb{R}^{c\times d_\beta}$ which are shared and learnable during training. Then, the input \textit{X} interact with the weights:
\begin{equation}
    (\textit{Q}, \textit{K}, \textit{V})=\textit{X}\cdot(\textit{W}_Q, \textit{W}_K, \textit{W}_V).
\end{equation}
Finally, the attention features \textit{Y} can be obtained by
\begin{equation}
    \textit{Y}=softmax(\frac{\textit{Q}\cdot{K}^T}{\sqrt{d_\alpha}})\textit{V}.
\end{equation}

In IR drop, the wider range of information around can also have a significant effect. Thus, it is necessary to construct a broader view within the attention mechanism. Feature models operating at the level of coarse-grained spatial grids identify and describe locations and relationships among global-scale structures.
However, it is still difficult to reduce false positive (FP) predictions for small objects that show large shape variations. In this paper, we demonstrate that higher accuracy can be achieved by integrating attention gates (AGs)\cite{oktay2018attention} into standard CNN models and utilizing self-attention to capture netlist embedding features. This approach does not require training multiple models or introducing a large number of additional model parameters. Additionally, cross-attention is employed to better fuse different modalities. Compared to multi-stage CNN models, attention modules gradually suppress the feature responses in irrelevant IR region, focusing instead on extracting the pressure drop information.

\subsection{Problem Formulation}
\label{sec:formulation}

\begin{mydefinition} [F$1$]
\label{def:f1}
To evaluate IR drop prediction accuracy, we employ a binary classification approach. Nodes with IR drop values exceeding 90\% of the maximum true value are classified as positive samples, while others are negative. 
Predictions are divided into four groups: True Positive (TP), True Negative (TN), False Positive (FP), and False Negative (FN).
The F1 score is calculated as follows:

\begin{equation}
    \mathrm{F1}=\frac{2 \times \mathrm{Precision} \times \mathrm{Recall}}{ \mathrm{Precision}+\mathrm{Recall}}.
\end{equation}

\begin{equation}
    \mathrm{Precision} = \frac{TP}{TP + FP},\quad \mathrm{Recall} = \frac{TP}{TP + FN}.
\end{equation}

F$1$ is used to assess the model's performance, particularly in identifying significant IR drop nodes. A higher F1 score indicates better accuracy in predicting critical IR drop.

\end{mydefinition}

\begin{mydefinition}[Mean absolute error (MAE)]
\label{def:mae}

In IR drop analysis, the accuracy of predicted voltage drops compared to actual measurements is often evaluated using the MAE. 
The metric is also used in the ICCAD 2023 contest. 
\begin{equation}
\text{MAE} = \frac{1}{n} \sum_{i=1}^{n} \left| V_{\text{pred}, i} - V_{\text{obs}, i} \right|.
\end{equation}

Lower MAE values indicate better predictive accuracy, as they reflect smaller deviations between predicted and observed voltages. IR-Drop predictor aiming to get lower MAE.

\end{mydefinition}

\begin{mydefinition}[Turn around time (TAT)]
\label{def:tat}
Inference time of the ML model. 
The goal is to have fast inference.
\end{mydefinition}


\begin{myproblem}
\label{def:ir-drop}
Given netlist files and 3 circuit-based features: current map, PDN density map, and effective distance map, which are detailed in \Cref{sec:circuit}, the objective is to train a multimodal-based framework that predicts the static IR drop within an acceptable inference time while achieving the highest F1 score, minimal mean absolute error (MAE). 
\end{myproblem}

\section{The Proposed IR Drop Prediction Flow}

\begin{figure*}[tb!]
    \centering
    \includegraphics[width=1.0\linewidth]{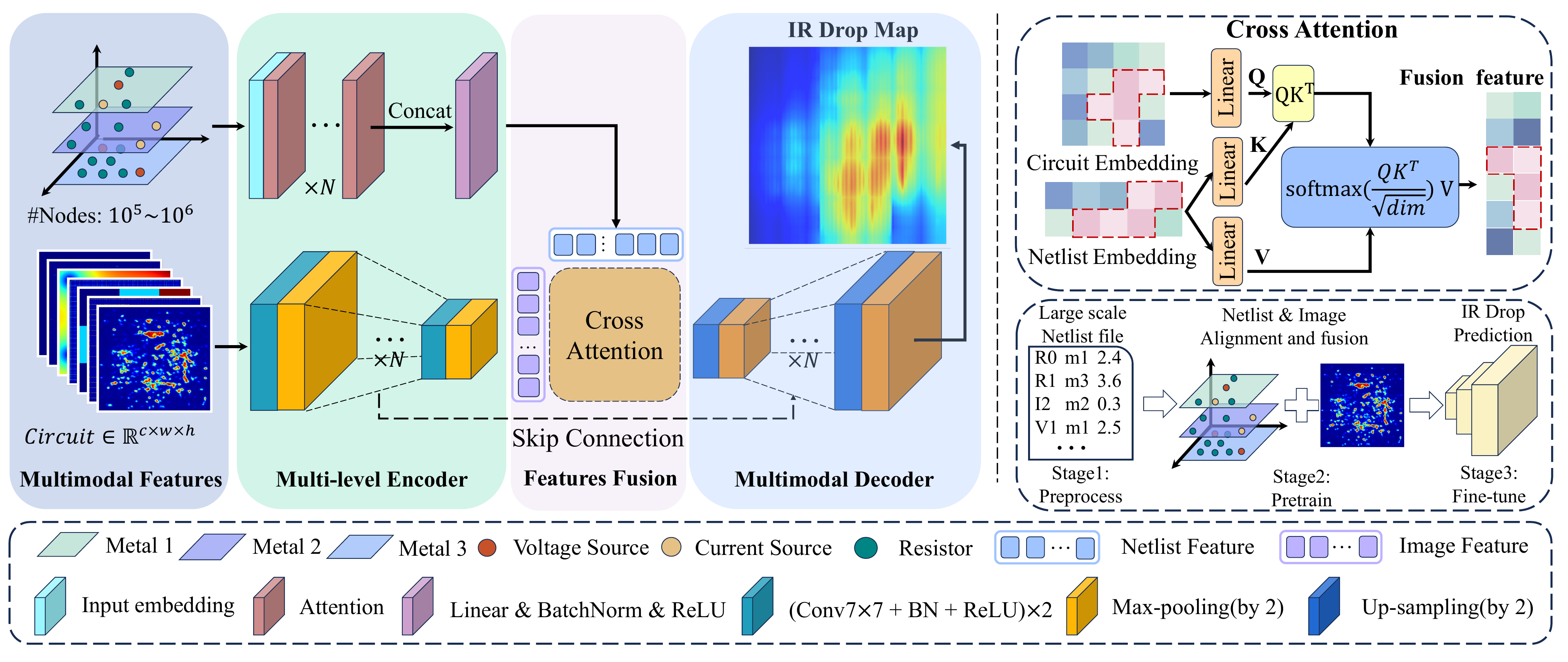}
    \caption{Our LMM-Net framework.
    }
    \label{fig:flow}
    \vspace{-1em}
\end{figure*}

\label{sec:ir-drop-flow}
As illustrated in \Cref{fig:flow}, we outline the complete architecture of our framework. We first introduce the circuit encoder. Subsequently, we present our large-scale netlist transformer (LNT) along with the direct netlist preprocessing approach. Then, our netlist encoding methodology is presented and offering insights in contrast to traditional netlist representation techniques. Lastly, we describe the fusion of the two modalities, followed by the multimodal decoder.

\subsection{Embedding PDN with Circuit Encoder}
\label{sec:enhance-dataset}

For circuit-based thermal data processing, we leverage downsampling encoder structure to effectively extract hierarchical spatial features. In order to capture a receptive field large enough to capture semantic context information, the feature map grid in the standard CNN structure is gradually downsampled.
With regard to the PDN structure-level features, this paper incorporates current, effective distance and PDN density map \cite{iccad23}. The effective distance is a metric that evaluates the distance from each location to all voltage sources. It is calculated as the reciprocal of the sum of the inverse Euclidean distances. Furthermore, the PDN density map is generated by extracting the mean PDN spacing within each grid \cite{began}. 
This paper also introduces voltage and current source plots and resistance maps to provide additional insight into the PDN structure. 
The voltage and current source plot are derived based on the values at their respective source positions, while the resistance map is generated by distributing the resistance of each resistor across the corresponding overlapping grid cells.

What's more, the samples in the dataset differed significantly in spatial dimension (width × height), with edge lengths ranging from $204$ to $930$. However, it is necessary to ensure that each batch has a consistent spatial dimension during model training. Therefore, it is crucial to employ spatial adjustment techniques. Considering the structure of the entire network, cropping the data may lead to information loss, which may affect the prediction. Eventually, in this paper, scaling and padding techniques are used to process the data in order to maximize the retention of overall information.

During the model training phase, each batch undergoes a scaling or padding operation. Padding is applied only when the edge length is less than 512 which allows lossless encoding; conversely, scaling is performed if the edge length exceeds 512. These spatial adjustment techniques ensure that the overall information is preserved and the stability of the data is maintained, thus improving the overall performance and robustness of the model. In addition, the corresponding normalization is performed for each channel. During model training, each batch of data is scaled or padded while the corresponding normalization process is performed for each channel. This normalization process helps to remove the inter-channel bias in the data, which in turn enhances the stability and generalization of the model. By mapping the range of values for each channel to similar intervals, normalization avoids the dominant influence of certain channels on model training, thus better preserving the statistical properties of the data. This further improves the model's ability to adapt to the input data and facilitates the model's generalization performance to unseen data.

\subsection{Embedding the netlist into point cloud}

Since convolutional neural networks (CNNs) are limited to capturing global information and pixel-based circuit representations often lack precision. To address this, we propose a novel large-scale netlist encoding approach that provides comprehensive global insights. As illustrated in \Cref{fig:netlist_embedding}, our approach can encode netlists with over 100k nodes without any information loss. In power delivery networks (PDNs), power delivery extends beyond a single layer, creating complex interactions that impact overall performance. Vias, which serve as vertical interconnects between different layers, play a critical role in enabling current flow across layers. However, this inter-layer connection introduces unique challenges, such as increased IR drops at via positions, which can degrade performance and signal integrity. Given their importance, accurately modeling and capturing the behavior of vias is essential for effective circuit analysis and optimization. 

Traditional netlist representations often convert netlists into $2$D maps. These methods obtain coordinate pairs ($x_1$, $y_1$) and ($x_2$, $y_2$) and place an average value between the start and end nodes. For nets spanning different layers, values are typically combined pixel-by-pixel. This approach ignores spatial connections. Unlike prior work, our approach encodes netlists using detailed attributes. These include coordinates ($x_1$, $y_1$), ($x_2$, $y_2$), values, element types (R, I, V), and key structural details such as originating and destination layers. If the layer$1$ is same as layer$2$, the component is a via. This rich and precise encoding captures node-wise connections and provides the model with a robust set of global features.

\subsection{Encoding netlist point cloud with LNT}
\label{sec:embedding-features}

For SPICE netlist processing, we first construct a 3D point cloud representation\cite{woo2002new} to capture the topological structure of the SoC. Each point in the cloud encodes critical physical and electrical characteristics including spatial coordinates ($x_1$, $y_1$, $x_2$, $y_2$) representing the node's position in the multi-layer structure, electrical parameters (node values), node type information, and layer identification.
To effectively encode SPICE-based features into a shared feature space with circuit, we leverage the Large-scale Netlist Transformer (LNT) architecture that is inspired by \cite{vaswani2017attention}. This approach significantly benefits from transformer-based architectural principles while extending them to 3D points data processing.

Unlike grid-based or image-based representations, point clouds naturally handle the irregular structure of circuit netlists, where components and connections do not follow a regular grid pattern. Point clouds efficiently represent large-scale netlists with millions of nodes without the memory overhead of dense matrix representations, making them particularly suitable for modern complex SoC designs\cite{7805857}. Point cloud representation enables seamless integration of both spatial information (3D coordinates) and electrical properties (node values, types) into a unified data structure, facilitating more effective feature learning.

\begin{figure}[tb!]
    \centering    \includegraphics[width=1.0\linewidth]{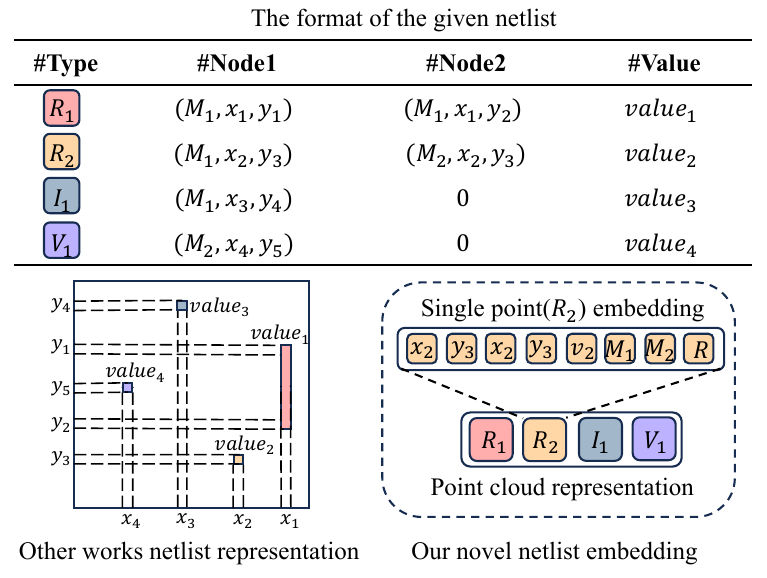}
    \caption{Illustration of ordinary netlist representation vs. our proposed netlist embedding.
    }
    \label{fig:netlist_embedding}
\end{figure}

The initial processing stage, illustrated in \Cref{fig:netlist_embedding}, transforms the circuit netlist into a point cloud representation. Each point in this representation encapsulates three key attributes: spatial coordinates, electrical values, and component classifications (including resistors, current sources, and voltage sources).

Subsequently, each netlist undergoes transformation into a high-dimensional vector space through a trainable embedding layer. These embeddings are optimized during the training phase to capture essential circuit characteristics and topological relationships inherent in the netlist structure.


\subsection{Multimodality Fusion Model for Static IR Drop}
\label{sec:model}


In the encoding phase, the circuit and netlist are encoded into features, respectively.
These multi-dimensional features are projected into a higher-dimensional embedding space through a learned transformation. The embedded features are then processed by an attention mechanism that learns to focus on relevant node interactions while suppressing less important ones. Finally, a fully-connected layer transforms the attention-weighted features into the desired output representation.

On the contrary, in the decoding stage, $4$ upsampling operations are performed, each with a coefficient of $2$, and each time deconvolution is used to recover the spatial dimension.
According to the observation, hotspots of static pressure drop are obviously not only affected by local information. 
Finally, the prediction result of the static pressure drop graph is obtained by $1\times1$ convolution operation.

Our framework adopts a dual-stream architecture that consists of four key components: circuit encoder, large-scale netlist transformer (LNT), multimodal fusion module, and decoder. The circuit encoder extracts spatial features from circuit layouts to capture physical power distribution patterns. In parallel, LNT processes netlist information through our direct netlist preprocessing approach, which encodes structural and connectivity features essential for IR drop analysis. The fusion module then combines spatial information from circuit maps and electrical characteristics from netlists, enabling comprehensive context awareness. Finally, the multimodal decoder leverages the fused representations to generate accurate IR drop.
This architecture is trained end-to-end with MSE loss that guides the model to learn critical IR drop while minimizing prediction errors. Our framework introduces an innovative two-stage training strategy. In the first phase, we employ a reconstruction task to train the network, enhancing the model's joint representation of circuit and netlist. Subsequently, the pre-trained model undergoes fine-tuning specifically for IR drop prediction. The trained model demonstrates strong generalization capabilities across different designs, offering fast and robust IR drop predictions by effectively capturing both spatial and electrical dependencies.




\section{Experiments}
\label{sec:experiments}

\begin{table}[tb!]
    \centering
    \caption{Statistics of the testcases.}
    \begin{threeparttable}
        \renewcommand{\arraystretch}{1.4}
        \resizebox{\columnwidth}{!}{%
        \begin{tabular}{lccccc}
            \hline
            \hline
            \textbf{Testcase} & \textbf{7} & \textbf{8} & \textbf{9} & \textbf{10} & \textbf{13} \\
            \hline
            Nodes & 85,591 & 83,030 & 166,734 & 159,940 & 15,768 \\
            Shape & 601 $\times$ 601 & 601 $\times$ 601 & 835 $\times$ 835 & 835 $\times$ 835 & 257 $\times$ 257 \\
            \hline
            \\
            \hline
            \hline
            \textbf{Testcase} & \textbf{14} & \textbf{15} & \textbf{16} & \textbf{19} & \textbf{20} \\
            \hline
            Nodes & 15,436 & 57,508 & 55,197 & 181,206 & 174,304 \\
            Shape & 257 $\times$ 257 & 489 $\times$ 489 & 489 $\times$ 489 & 870 $\times$ 870 & 870 $\times$ 870 \\
            \hline
        \end{tabular}%
        }
        \begin{tablenotes}
            \footnotesize
            \item Shape is measured in pixels.
        \end{tablenotes}
    \end{threeparttable}
    \label{tab:testcase}
\end{table}

\begin{table*}[ht]
  \centering
  \caption{Comparison with state of the arts.}
  \label{tab:results}
  \resizebox{2.0\columnwidth}{!}{
  \begin{tabular}{*{16}{c}}
    \toprule
    \multirow{2}*{Circuits} & \multicolumn{3}{c}{1st Place \cite{iccad23}} & \multicolumn{3}{c}{2nd Place \cite{iccad23}} & \multicolumn{3}{c}{IREDGe \cite{began}} & \multicolumn{3}{c}{IRPnet \cite{date24}} & \multicolumn{3}{c}{Ours} \\
    \cmidrule(lr){2-4}\cmidrule(lr){5-7}\cmidrule(lr){8-10}\cmidrule(lr){11-13}\cmidrule(lr){14-16} & F$1$ & MAE & TAT & F$1$ & MAE & TAT & F$1$ & MAE & TAT & F$1$ & MAE & TAT & F$1$ & MAE & TAT\\
    \midrule
    testcase$7$ & $\mathbf{0.78}$ & $0.66$ & $14.61$ & $0.56$ & $0.78$ & $3.22$ & $0.16$ & $5.77$ & $\mathbf{1.53}$ & $0.17$ & $2.39$ & $2.87$ & $0.72$ & $\mathbf{0.63}$ & $2.82$ \\
    testcase$8$ & $0.82$ & $\mathbf{0.82}$ & $12.64$ & $0.8$ & $1.13$ & $2.70$ & $0.20$ & $4.20$ & $\mathbf{1.27}$ & $0.10$ & $2.30$ & $2.43$ & $\mathbf{0.84}$ & $0.84$ & $2.57$ \\
    testcase$9$ & $\mathbf{0.59}$ & $\mathbf{0.41}$ & $18.84$ & $0.55$ & $0.73$ & $4.25$ & $0.04$ & $4.71$ & $\mathbf{2.42}$ & $0.00$ & $5.05$ & $3.46$ & $0.47$ & $0.42$ & $4.63$ \\
    testcase$10$ & $0.53$ & $\mathbf{0.66}$ & $19.05$ & $0.15$ & $1.14$ & $\mathbf{4.13}$ & $0.01$ & $4.76$ & $2.67$ & $0.00$ & $2.02$ & $2.89$ & $\mathbf{0.60}$ & $0.71$ & $4.43$ \\
    testcase$13$ & $0.00$ & $2.07$ & $9.60$ & $\mathbf{0.67}$ & $\mathbf{1.25}$ & $1.25$ & $0.38$ & $8.42$ & $1.64$ & $0.01$ & $5.78$ & $1.22$ & $0.52$ & $1.52$ & $\mathbf{1.15}$ \\
    testcase$14$ & $0.00$ & $4.22$ & $10.07$ & $0.10$ & $\mathbf{2.32}$ & $1.40$ & $0.05$ & $7.43$ & $1.99$ & $0.00$ & $2.33$ & $1.13$ & $\mathbf{0.44}$ & $3.24$ & $\mathbf{1.11}$ \\
    testcase$15$ & $0.09$ & $\mathbf{0.97}$ & $12.99$ & $0.00$ & $1.92$ & $2.15$ & $0.1$ & $5.48$ & $\mathbf{1.77}$ & $0.00$ & $5.51$ & $2.88$ & $\mathbf{0.54}$ & $1.49$ & $2.20$ \\
    testcase$16$ & $0.53$ & $\mathbf{1.60}$ & $12.12$ & $0.48$ & $3.44$ & $2.19$ & $0.31$ & $10.21$ & $\mathbf{0.97}$ & $0.01$ & $5.78$ & $2.21$ & $\mathbf{0.55}$ & $3.33$ & $2.43$ \\
    testcase$19$ & $0.50$ & $0.91$ & $19.05$ & $0.49$ & $1.20$ & $4.55$ & $0.05$ & $4.62$ & $\mathbf{2.52}$ & $0.01$ & $2.71$ & $3.14$ & $\mathbf{0.61}$ & $\mathbf{0.74}$ & $4.60$ \\
    testcase$20$ & $\mathbf{0.71}$ & $1.18$ & $18.75$ & $0.74$ & $1.07$ & $4.58$ & $0.02$ & $7.24$ & $3.39$ & $0.00$ & $5.91$ & $\mathbf{3.12}$ & $0.54$ & $\mathbf{0.64}$ & $4.61$ \\
    \midrule
    $Avg$ & $0.46$ & $1.35$ & $14.77$ & $0.45$ & $1.50$ & $3.04$ & $0.13$ & $6.28$ & $\mathbf{2.02}$ & 
    $0.03$ & $3.98$ & $2.54$ & $\mathbf{0.58}$ & $\mathbf{1.35}$ & $3.05$ \\
    $Ratio$ & $0.80$ & $1.00$ & $4.84$ & $0.77$ & $1.11$ & $0.99$ & $0.22$ & $4.65$ & $\mathbf{0.66}$ & $0.05$ & $2.95$ & $0.83$ & $\mathbf{1.00}$ & $\mathbf{1.00}$ & $1.00$ \\
    \bottomrule
    \multicolumn{16}{l}{\footnotesize{MAE in $10^{-4}$, TAT in $second$.}} \\
\end{tabular}
  }
\end{table*}

\subsection{Experiments Settings and Benchmarks}
\label{sec:experiments-settings}

We test our Predictor on a platform with NVIDIA GeForce RTX H100 80GB GPU. 
The programming environment is Python 3.9.18, and the PyTorch version is 2.1.1 with the CUDA version 12.3.

We select $512 \times 512$ as the input side length for each batch and use \texttt{Adam} as the optimizer, with a batch size of 16. 
The initial learning rate is set as $0.001$ with a total epoch of $200$. 
The ICCAD2023 contest \cite{iccad23} provided $100$ fake cases and 10 real cases for training, plus $10$ real cases for testing. 
Due to limited training data, $2000$ cases provided by BeGAN \cite{began} are used to expand the dataset in this paper. 
In addition, since the data provided by the ICCAD2023 competition is very similar to the distribution of real cases during the test phase, this paper over-samples each fake case $10$ times and each real case $20$ times. Therefore, the training set contains a total of $3310$ cases. Finally, we test our model on the $10$ hidden case shown in \Cref{tab:testcase}. 

\subsection{Comparisons with SOTA works}
\label{sec:results}

In \Cref{tab:results}, we compare our proposed algorithm against the winning team of the ICCAD2023 competition, as well as the IRPnet \cite{date24} and IREDGe \cite{began-net} methods, which are re-implemented in PyTorch for this paper. 
It is worth noting that the third-party code we tested experiences runtime issues and crashes in the last two cases. Despite this, the experimental results show that our algorithm's F1 score improves by 20\% compared to the first-place team. This is noteworthy, as the second-place team enhanced their training data by generating additional cases with parameter modifications (around 5,400 cases in total), giving their model a significant competitive advantage.
Our algorithm achieves the best F1 score in 6 out of the tested cases, as well as the overall best average F1 score. In contrast, IREDGe \cite{began-net} performs poorly due to its limited feature set, which includes only the current map, PDN density map, and valid distance map, as well as its simpler models. IRPnet \cite{date24} was only tested on ten real-circuit data samples, so we reconstructed it to test on the hidden-circuit data. In the hidden case, there are several cases that differs from the training set. The results show that IRPnet does not demonstrate sufficient generalization to the hidden cases.

\begin{figure}[tb!]
    \vspace{-1em}
    \centering    \includegraphics[width=1.0\linewidth]{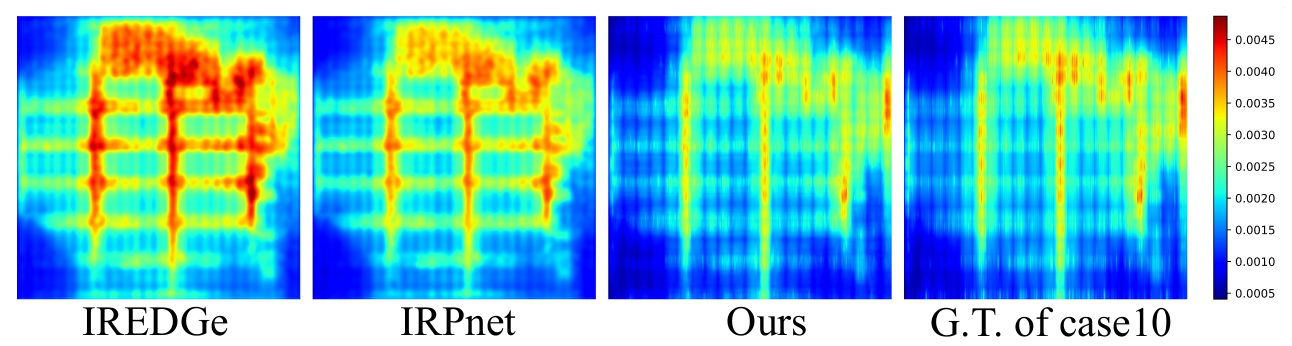}
    \caption{The IR drop prediction visualizations of SOTA works and ours.
    }
    \label{fig:result}
\end{figure}

In conclusion, our proposed algorithm achieves excellent and robust performance without incurring a significant time overhead. \Cref{fig:result} provides a visual comparison of the circuit IR Drop map generated by IREDGe, IRPnet, our method, and ground truth on the testcase$10$.

\subsection{Ablation studies}

\begin{figure}[t]
  \vspace{-1em}
  \centering
  \pgfplotsset{
    width=0.95\linewidth,
    height=0.5\linewidth
}

\begin{tikzpicture}[scale=1.0]
    \begin{axis}[
        name=plot1,
        ybar,
        xticklabels={EC, W-Att, W-LNT, W-Aug, United},
        xtick={1.2,2.2,3.2,4.2,5.2},
        xtick align=inside,
        xticklabel style={font=\scriptsize},
        ylabel={\scriptsize F1},
        ylabel style={font=\scriptsize},
        ytick={0.2, 0.4, 0.6, 0.8},
        yticklabel style={font=\scriptsize},
        ylabel near ticks,
        bar width=10pt,
        xmin=0.4,
        xmax=5.9,
        ymin=0,
        ymax=0.8,
        nodes near coords,
        nodes near coords style={
            font=\scriptsize\fontfamily{ptm}\selectfont,
            /pgf/number format/.cd,
            fixed zerofill,
            precision=2
        },
        legend style={
            font=\scriptsize,
            at={(0.128,0.99)},
            draw=none,
            anchor=north,
            legend columns=-1
        },
        legend entries={F1}
    ]
    \addplot[ybar, fill=myblue, draw=black, area legend] coordinates {
        (1, 0.27) (2, 0.30) (3, 0.48) (4, 0.13) (5, 0.58)
    };
    \end{axis}
    
    \begin{axis}[
        name=plot2, 
        ybar,
        hide x axis,
        axis y line*=right,
        ylabel={MAE ($\times 10^{-4}$)},
        ylabel style={font=\scriptsize},
        ytick={1.0, 2.0, 3.0, 4.0},
        yticklabel style={font=\scriptsize},
        bar width=10pt,
        xmin=0.05,
        xmax=5.55,
        ymin=0,
        ymax=4.0,
        nodes near coords,
        nodes near coords style={
            font=\scriptsize\fontfamily{ptm}\selectfont,
            /pgf/number format/.cd,
            fixed zerofill,
            precision=2
        },
        legend style={
            font=\scriptsize,
            at={(0.15,0.85)},
            draw=none,
            anchor=north,
            legend columns=-1
        },
        legend entries={MAE},
        at={(plot1.south west)},
        anchor=south west       
    ]
    \addplot[ybar, fill=myorange, draw=black, area legend] coordinates {
        (1, 1.93) (2, 2.65) (3, 1.96) (4, 2.03) (5, 1.35)
    };
    \end{axis}
\end{tikzpicture}
  \caption{Ablation studies on ICCAD-2023 contest dataset for different techniques' application.}
  \label{fig:ablation}
\end{figure}
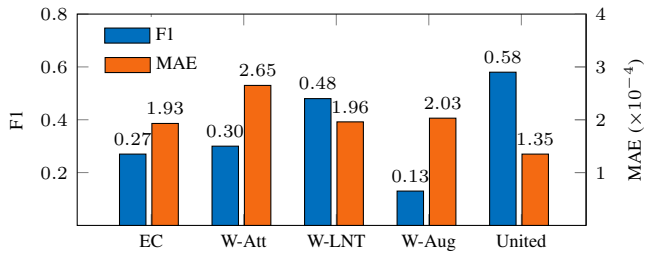

As shown in \Cref{sec:embedding-features}, we use a netlist transformer to process large-scale netlists. The ordinary encoder-decoder architecture often struggles to capture the global information that can represent the intricate circuit connections. To address this challenge, we leverage the self-attention block to focus on the high IR-Drop areas, which are a significant factor in the prediction. Additionally, due to the limited data available and the inherent circuit connection relationships within the data, simple cropping and flipping could potentially disrupt the circuit characteristics. Therefore, we employ Gaussian noise augmentation with a standard deviation range of $(0, 1e^{-3})$ to enrich our training set.

To evaluate the effectiveness of the techniques mentioned above, we conduct experiments on the ICCAD2023 benchmark. The results are shown in \Cref{fig:ablation}. The different configurations tested are: EC (only encoder-decoder flow), W-Att (without attention block), W-Aug (without augmentation), W-LNT (without large netlist transformer), and United (using all techniques together). 
Based on the results shown in \Cref{fig:ablation}, we can evaluate the impact of each technique. For F$1$ scores, the techniques are ranked in terms of impact from highest to lowest as follows: LNT, attention mechanism, and data augmentation. In terms of MAE, the impact order is LNT, data augmentation, and attention mechanism. The large-scale netlist transformer (LNT) technique has the most significant impact, representing a key contribution to our approach for hierarchical netlist encoding.
These results highlight the importance of hierarchical netlist encoding and the effectiveness of the proposed techniques in improving the model's performance on the ICCAD2023 contest benchmark.

\section{Discuss and Conclusion}
\label{sec:conclu}

In this paper, a Multimodal Fusion Model based on LNT embedding large-scale netlist is firstly proposed for the task of static voltage drop prediction. Meanwhile, additional feature maps are introduced to enhance the model's learning ability. Experimental results show that the algorithm proposed in this paper achieves the best F$1$ score and minimun mae compared with the winning team of ICCAD $2023$ competition and the most advanced method currently. Future research will further explore the application of specific large language model (LLM) to address large text information in PDN structures to better optimize the model.

\section*{Acknowledgment}

This work is supported by the National Natural Science Foundation of China (NSFC) under grant No. 92473113.


{
    \bibliographystyle{IEEEtran}
    \bibliography{ref/Top,ref/intro,ref/algo,ref/result, ref/flow, ref/prelim}
}

\end{document}